\documentclass[letterpaper]{article} 
\usepackage{aaai23}  
\usepackage{times}  
\usepackage{helvet}  
\usepackage{courier}  
\usepackage[hyphens]{url}  
\usepackage{graphicx} 
\usepackage{natbib}  
\usepackage{caption} 
\usepackage{algorithm}

\usepackage{newfloat}
\usepackage{listings}
\DeclareCaptionStyle{ruled}{labelfont=normalfont,labelsep=colon,strut=off} 
\floatstyle{ruled}
\newfloat{listing}{tb}{lst}{}
\floatname{listing}{Listing}
\usepackage{cite}
\usepackage{amsmath,amssymb,amsfonts}
\usepackage{graphicx}
\usepackage{textcomp}
\usepackage{xcolor}

\usepackage{soul}

\usepackage{amsthm}
\usepackage{algorithm}
\usepackage{algorithmicx}
\usepackage{algpseudocode}
\usepackage{caption}
\usepackage{subcaption}
\usepackage{multirow}
\usepackage{color}
\usepackage{balance}
\usepackage{enumitem}
\usepackage[normalem]{ulem}
\usepackage{booktabs}
\usepackage{wrapfig}
\usepackage[figuresright]{rotating}

\newtheorem{theorem}{Theorem}
\newtheorem{lemma}{Lemma}

\newtheorem{definition}{Definition}

\newcommand{\bL}{\ensuremath{\mathcal{L}}}

\newcommand{\bS}{\ensuremath{\mathcal{S}}}

\newcommand{\bG}{\ensuremath{\mathcal{G}}}
\newcommand{\bN}{\ensuremath{\mathcal{N}}}
\newcommand{\bE}{\ensuremath{\mathcal{E}}}
\newcommand{\bT}{\ensuremath{\mathcal{T}}}

\newcommand{\bM}{\ensuremath{\mathcal{M}}}

\newcommand{\bQ}{\ensuremath{\mathcal{Q}}}

\renewcommand{\vec}[1]{\ensuremath{\mathbf{#1}}}
\newcommand{\stitle}[1]{\vspace{2mm} \noindent {\bf #1}}
\newcommand{\eg}{{\it e.g.}}

\newcommand{\ie}{{\it i.e.}}
\newcommand{\etc}{{\it etc.}}
\newcommand{\wrt}{w.r.t. }

\newcommand{\method}[1]{#1}
\newcommand{\model}{\method{Count-GNN}{}}
\newcommand{\eat}[1]{}

\newcommand\zemin[1]{\textcolor{blue}{#1}}

\newcommand\iclrcite[1]{\cite{#1}}
\newcommand\iclrref[1]{(\ref{#1})}

\newcommand\kdd[1]{{#1}}

\def\equalcontribNew{%
      \ifnum\value{eqfn}=0%
        \footnote{Co-first authors with equal contribution. Part of the work was done while at Singapore Management University.}%
        \setcounter{eqfn}{\value{footnote}}%
      \else%
        \footnotemark[\value{eqfn}]%
      \fi%
    }%

\def\corresAuthor{%
      \ifnum\value{eqfn}=1%
        \footnote{Corresponding authors.}%
        \setcounter{eqfn}{\value{footnote}}%
      \else%
        \footnotemark[\value{eqfn}]%
      \fi%
    }%

\title{Learning to Count Isomorphisms with Graph Neural Networks}

\author{
Xingtong Yu, \textsuperscript{\rm 1}\equalcontribNew\
Zemin Liu, \textsuperscript{\rm 2}\equalcontribNew\
Yuan Fang, \textsuperscript{\rm 3}\corresAuthor\
Xinming Zhang \textsuperscript{\rm 1}\corresAuthor\
}
\affiliations{
    \textsuperscript{\rm 1} University of Science and Technology of China, China\\
    \textsuperscript{\rm 2} National University of Singapore, Singapore\\
    \textsuperscript{\rm 3} Singapore Management University, Singapore\\
    yxt95@mail.ustc.edu.cn, zeminliu@nus.edu.sg, yfang@smu.edu.sg, xinming@ustc.edu.cn
}

\begin{document}
\maketitle
\begin{abstract}
Subgraph isomorphism counting is an important problem on graphs, 
as many graph-based tasks exploit recurring subgraph patterns. Classical methods usually boil down to a backtracking framework that needs to navigate a huge search space with prohibitive computational costs. 
Some recent studies resort to graph neural networks (GNNs) to learn a low-dimensional representation for both the query and input graphs, in order to predict the number of subgraph isomorphisms on the input graph. However, typical GNNs employ a node-centric message passing scheme that receives and aggregates messages on nodes,
which is inadequate in complex structure matching for isomorphism counting. Moreover, on an input graph, the space of possible query graphs is enormous, and 
different parts of the input graph will be triggered to match different queries. Thus, expecting a fixed representation of the input graph to match diversely structured query graphs is unrealistic. 
In this paper, we propose a novel GNN called \model\ for subgraph isomorphism counting, to deal with the above challenges. At the edge level, given that an edge is an atomic unit of encoding graph structures, we propose an \emph{edge-centric message passing} scheme, where messages on edges are propagated and aggregated based on the edge adjacency to preserve fine-grained structural information.
At the graph level, we \emph{modulate the input graph representation} conditioned on the query, so that the input graph can be adapted to each query individually to improve their matching. Finally, we conduct extensive experiments on a number of benchmark datasets to demonstrate the superior performance of \model. 
\end{abstract}

\section{Introduction}
Research in network science and graph mining often finds and exploits recurring subgraph patterns on an input graph. 
For example, on a protein network, we could query for the hydroxy groups which consist of one oxygen atom covalently bonded to one hydrogen atom; on a social network, we could query for potential families in which several users form a clique and two of them are working and the rest are studying. These queries essentially describe a subgraph pattern that repeatedly occurs on different parts of an input graph, which expresses certain semantics such as the hydroxy groups or families. These subgraph patterns are also known as network motifs on homogeneous graphs \iclrcite{milo2002network} or meta-structures on heterogeneous graphs \iclrcite{sun2011pathsim,fang2016semantic}.
To leverage their expressiveness, more sophisticated graph models \iclrcite{monti2018motifnet,liu2018subgraph,sankar2019meta,wang2019heterogeneous} have also been designed to incorporate motifs or meta-structures.

The need for subgraph patterns in graph-based tasks and models leads to a high demand of \emph{subgraph isomorphism counting} \iclrcite{liu2020neural}. Classical methods usually employ search-based algorithms such as backtracking \iclrcite{ullmann1976algorithm,cordella2004sub,he2008graphs} to exhaustively detect the isomorphisms and return an exact count. However, their computational costs are often excessive given that  the detection problem is NP-complete and the counting form is \#P-complete \cite{cordella2004sub}. With the rise of graph neural networks (GNNs) \iclrcite{wu2020comprehensive}, some recent approaches for subgraph isomorphism counting also leverage on the powerful graph representations from GNNs \iclrcite{liu2020neural,zhengdao2020can,xia2022substructure}. They generally employ GNNs to embed the queries and input graphs into low-dimensional vectors, which are further fed into a counter module to predict the approximate number of isomorphisms on the input graph.
Compared to classical approaches, they can significantly save computational resources at the expense of  approximation, providing a useful trade-off between accuracy and cost since many applications do not necessarily need an exact count. 

However, previous GNN-based isomorphism counting models adopt a node-centric message-passing scheme, which propagates and aggregates messages on nodes. While this scheme is effective for node-oriented tasks, it falls short of matching complex structures for isomorphism counting.
In particular, they  rely on message aggregation to generate representations centering on nodes, failing to explicitly and fundamentally capture the  complex interactions  among nodes. 
Thus, as the first challenge, \emph{how do we capture fine-grained structural information} beyond node-centric GNNs?
Moreover, on an input graph, the space of possible query graphs is enormous. Different queries are often characterized by distinct structures that match with different parts of the input graph. A fixed graph representation to match with all possible queries is likely to underperform. Thus, as the second challenge, \emph{how do we adapt the input graph to each query individually}, in order to improve the matching of specific structures in every query?

\begin{figure}[t]
\centering
\includegraphics[width=1\linewidth]{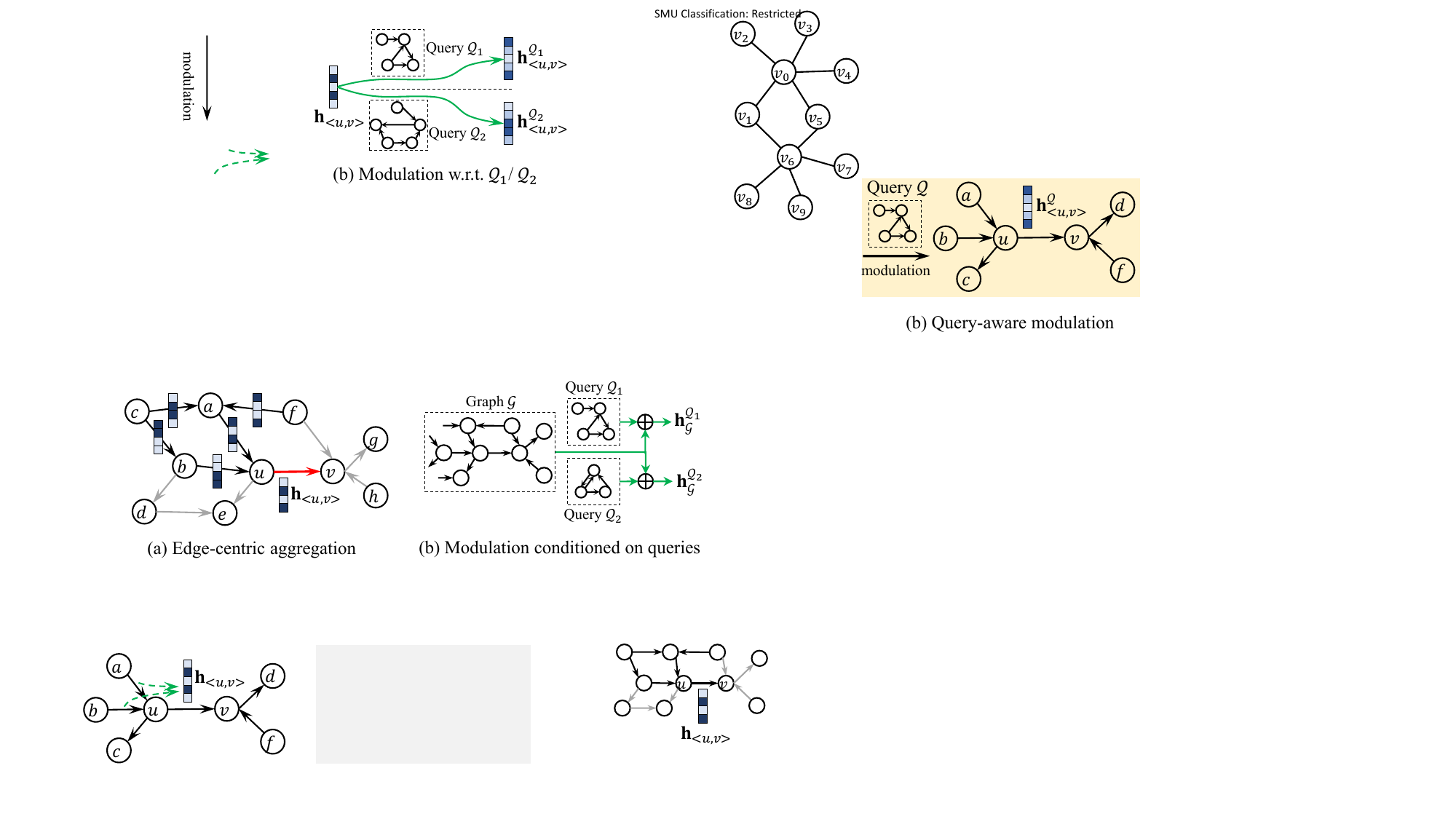}
\vspace{-3mm}
\caption{Illustration of \model.}
\label{fig.dual-views}
\vspace{-1mm} 
\end{figure}

In this paper, we propose a novel model called \model\ for approximate subgraph isomorphism counting, which copes with the above challenges from both the edge and graph perspectives.
To be more specific, at the edge level, \model\ is built upon an \emph{edge-centric} GNN that propagates and aggregates messages on and for edges based on the edge adjacency, as shown in Fig.~\ref{fig.dual-views}(a). 
Given that edges constitute the atomic unit of graph structures, any subgraph is composed of one or more edge chains. Thus, treating edges as first-class citizens can better capture fine-grained structural information.
Theoretically, our proposed edge-centric GNN model can be regarded as a generalization of node-centric GNNs, with provably stronger expressive power than node-centric GNNs.
At the graph level, \model\ resorts to a modulation mechanism \iclrcite{perez2018film} by adapting the input graph representation 
to each query graph, as shown in Fig.~\ref{fig.dual-views}(b). As a result, the input graph can be tailored to each individual query with varying structures. Coupling the two perspectives, \model\ is able to precisely match complex structures between the input graph and structurally diverse queries. 

To summarize, our contributions are three-fold.
(1) We propose a novel model \model\ that capitalizes on edge-centric aggregation to encode fine-grained structural information, 
which is more expressive than node-centric aggregation in theory.
(2) Moreover, we design a  query-conditioned graph modulation in \model, to adapt structure matching to different queries from the graph perspective.
(3) Extensive experiments on several benchmark datasets demonstrate that \model\ can significantly outperform state-of-the-art GNN-based models for  isomorphism counting.

\section{Related Work}
We present the most related studies here, while leaving the rest to Appendix~G due to the space limitation.

Graphs usually entail abundant local structures to depict particular semantics, which gives rise to the importance of subgraph isomorphism counting \cite{ullmann1976algorithm}. 
To solve this problem, most traditional methods resort to backtracking \cite{ullmann1976algorithm,cordella2004sub,he2008graphs}. Although they can obtain the exact counts, the search space usually grows intractably as graph size increases. In fact, subgraph isomorphism counting is a \#P-complete problem \cite{ullmann1976algorithm}.
Subsequently, several approaches \cite{han2013turboiso,carletti2017challenging} are proposed to utilize some constraints to reduce the search space, and others \cite{yan2004graph} try to filter out infeasible graphs to speed up the backtracking process. 
Another line of approaches \cite{alon1995color,bressan2021faster} rely on the color coding for subgraph isomorphism counting in polynomial time. They are usually fixed-parameter tractable and can only be employed for some limited subcases.
Other studies perform count estimation \cite{teixeira2020sequential,pinar2017escape,teixeira2018graph,wang2014efficiently}, such as in the setting where access to the entire network is prohibitively expensive \cite{teixeira2020sequential}, or using the counts of smaller patterns to estimate the counts for larger ones \cite{pinar2017escape}. However, these attempts still face high computational costs.

Recently, a few studies \cite{liu2020neural,zhengdao2020can} propose to address subgraph isomorphism counting from the perspective of machining learning. 
One study \cite{liu2020neural} proposes to incorporate several existing pattern extraction mechanisms such as CNN \cite{lecun1998gradient}, GRU \cite{chung2014empirical} and GNNs \cite{wu2020comprehensive} on both query graphs and input graphs to exploit their structural match, which is then followed by a counter module to predict the number of isomorphisms. 
Another work \cite{zhengdao2020can} analyzes the ability of GNNs in detecting subgraph isomorphism, and proposes a Local Relational Pooling model based on the permutations of walks according to breadth-first search to count certain queries on graphs. However, they need to learn a new model for each query graph, limiting their practical application.
Compared to traditional methods, these machine learning-based models can usually approximate the counts reasonably well, and at the same time significantly save computational resources and time, providing a practical trade-off between accuracy and cost. However, these approaches only adopt node-centric GNNs, which limit their ability to capture finer-grained structural information.

There also exist a few recent GNNs based on edge-centric aggregation \cite{monti2018dual,jiang2020co}, node-centric aggregation with the assistance of edge features \cite{gong2019exploiting,yang2020nenn,isufi2021edgenets}, or both node- and edge-centric aggregation at the same time \cite{liu2022graph}.
Except DMPNN \cite{liu2022graph}, they are not specifically devised for subgraph isomorphism counting. In particular, they all lack the key module of structural matching between query and input graphs. 
Besides, the edge-centric approaches do not theoretically justify their enhanced expressive power compared to the node-centric counterparts.

\section{Problem Formulation}
A \emph{graph} $\bG=(V_\bG, E_\bG)$ is defined by a set of nodes $V_\bG$, and a set of edges $E_\bG$ between the nodes. In our study, we consider the general case of directed edges, where an undirected edge can be treated as two directed edges in opposite directions. 
We further consider \emph{labeled} graphs (also known as heterogeneous graphs), in which there exists a node label function $\ell: V_\bG\rightarrow L$  and an edge label function $\ell': E_\bG\rightarrow L'$, where $L$ and $L'$ denote the set of labels on nodes and edges, respectively. 
A graph $\bS=(V_\bS, E_\bS)$ is a \emph{subgraph} of $\bG$, written as $\bS \subseteq \bG$, if and only if $V_\bS\subseteq V_\bG$ and $E_\bS\subseteq E_\bG$.

Next, we present the definition of subgraph isomorphism on a labeled graph. 
\begin{definition}[Labeled Subgraph Isomorphism]\label{def.isomorphic}
Consider a subgraph $\bS$ of some \emph{input graph}, and a \emph{query graph} $\bQ$. $\bS$ is \emph{isomorphic} to $\bQ$, written as $\bS \simeq \bQ$, if there exists a bijection between their nodes, $\psi: V_{\bS}\rightarrow V_{\bQ}$, such that 
\begin{itemize}
    \item $\forall v\in V_{\bS}$, $\ell(v)=\ell(\psi(v))$;
    \item $\forall e=\langle u,v\rangle \in E_{\bS}$, it must hold that $e'=\langle \psi(u),\psi(v)\rangle \in E_{\bQ}$ and
$\ell'(e)=\ell'(e')$.  \qed
\end{itemize}
\end{definition}

\begin{figure*}[t]
\centering
\includegraphics[width=0.85\linewidth]{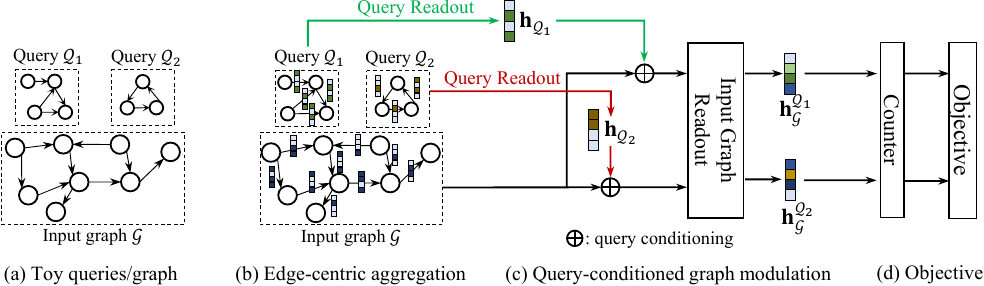}
\caption{Overall framework of \model.}
\label{fig.framework}
\vspace{-1mm} 
\end{figure*}

In the problem of \emph{subgraph isomorphism counting}, we are given a query graph $\bQ$ and an input graph $\bG$. We aim to predict $n(\bQ,\bG)$, the number of subgraphs on $\bG$ which are isomorphic to $\bQ$, \ie, the cardinality of the set $\{\bS  | \bS\subseteq \bG,\bS \simeq \bQ \}$.
Note that this is a non-trivial \#P-complete problem \iclrcite{cordella2004sub}. In practice, the query $\bQ$ usually has a much smaller size than the input graph $\bG$, \ie, $|V_{\bQ}|\ll|V_{\bG}|$ and $|E_{\bQ}|\ll|E_{\bG}|$, leading to a huge search space and computational cost.

\section{Proposed Model: \model}
In this section, we  present the overall framework of \model\ first, followed by individual modules.
\subsection{Overall Framework}
We give an overview of the proposed \model\ in Fig.~\ref{fig.framework}. Consider some query graphs and an input graph in Fig.~\ref{fig.framework}(a). On both the query and input graphs, we first conduct edge-centric aggregation in which messages on edges are propagated to and aggregated for each edge based on the edge adjacency, as shown in Fig.~\ref{fig.framework}(b). This module targets at the edge level, and enables us to learn edge-centric representations for both input graphs and queries that capture their fine-grained structural information for better structure matching. Furthermore, to be able to match diverse queries with distinct structures, the edge  representations of the input graph are modulated w.r.t.~each query, as shown in Fig.~\ref{fig.framework}(c). The query-conditioned edge representations then undergo a readout operation to fuse into a query-conditioned whole-graph representation for the input graph. The module targets at the graph level, and enables us to adapt the input graph to 
each query individually to improve the matching of specific structures in each query. 
Finally, as shown in Fig.~\ref{fig.framework}(d), a counter module is applied to predict the isomorphism counting on the input graph for a particular query, forming the overall objective.

\subsection{Edge-Centric Aggregation} \label{sec.edge-centric}

Typical GNNs \iclrcite{kipf2016semi,velivckovic2017graph,hamilton2017inductive} and GNN-based isomorphism counting models \iclrcite{liu2020neural,zhengdao2020can}  resort to the key mechanism of node-centric message passing, in which each node receives and aggregates messages from its neighboring nodes. 
For the problem of subgraph isomorphism counting, it is crucial to capture fine-grained structural information for more precise structure matching between the query and input graph. Consequently, we exploit edge-centric message passing, in which each edge receives and aggregates messages from adjacent edges. The edge-centric GNN captures structural information in an explicit and fundamental manner, given that edges represent the atomic unit of graph structures.

More concretely, we learn a representation vector for each edge by propagating messages on edges. A message can be an input feature vector of the edge in the input layer of the GNN, or an intermediate embedding vector of the edge in subsequent layers.
Specifically, given a directed edge $e=\langle u, v \rangle$ on a graph (either an input or query graph), we initialize its message as a $d_0$-dimensional vector
\begin{align}
    \mathbf{h}_{\langle u, v \rangle}^0=\mathbf{x}_u \parallel \mathbf{x}_{\langle u, v \rangle} \parallel \mathbf{x}_v\in \mathbb{R}^{d_0},
\end{align} 
where $\mathbf{x}_*$ encodes the input features of the corresponding nodes or edges and $\parallel$ is the concatenation operator. In general, $\mathbf{h}_{\langle u, v \rangle}^0 \ne \mathbf{h}_{\langle v, u \rangle}^0$ for directed edges. Note that, in the absence of input features, we can employ one-hot encoding as the feature vector; it is also possible to employ additional embedding layers to further transform the input features into initial messages.

Given the initial messages, we devise an edge-centric GNN layer, in which each edge receives and aggregates messages along the directed edges. The edge-centric message passing can be made recursive by stacking multiple layers. Formally, in the $l$-th layer, the message on a directed edge ${\langle u,v\rangle}$, \ie, $\vec{h}^l_{\langle u,v\rangle}\in\mathbb{R}^{d_l}$, is updated as
\begin{equation} \label{eq.egnn}
    \vec{h}^l_{\langle u,v\rangle} = \sigma(\vec{W}^l\vec{h}^{l-1}_{\langle u,v\rangle} + \vec{U}^l{\vec{h}}^{l-1}_{\langle \cdot,u\rangle} + \vec{b}^l),
\end{equation}
where $\vec{W}^l$, $\vec{U}^l\in\mathbb{R}^{d_l\times d_{l-1}}$ are learnable weight matrices, $\vec{b}^l\in\mathbb{R}^{d_l}$ is a learnable bias vector, and $\sigma$ is an activation function (we use LeakyReLU in the implementation). 
In addition, ${\vec{h}}^{l-1}_{\langle \cdot,u\rangle}\in\mathbb{R}^{d_{l-1}}$ is the intermediate message aggregated from the preceding edges of $\langle u,v\rangle$, \ie, edges incident on node $u$ from other nodes, which can be materialized as
\begin{equation} \label{eq.aggr}
    {\vec{h}}^{l-1}_{\langle \cdot,u\rangle} = \textsc{Aggr}(\{\vec{h}^{l-1}_{\langle i,u\rangle}| \langle i,u\rangle\in E\}),
\end{equation}
where $E$ denotes the set of directed edges in the graph, and $\textsc{Aggr}(\cdot)$ is an aggregation operator to aggregate messages from the preceding edges. We implement the aggregation operator as a simple mean, although more sophisticated approaches such as self-attention \iclrcite{hamilton2017inductive} and multi-layer perceptron \iclrcite{xu2018powerful} can also be employed. 
To boost the message passing, more advanced mechanisms can be imported into the layer-wise edge-centric aggregation, \eg, a residual \iclrcite{he2016deep} can be added to assist the message passing from previous layers to the current layer.

The above multi-layer edge-centric aggregation is applied to each query and input graph in a dataset. 
All query graphs share one set of GNN parameters (\ie, $\vec{W}^l, \vec{U}^l, \vec{b}^l$), while all input graphs share another set.
On all graphs, the aggregated message on an edge \kdd{$e={\langle u, v \rangle}$} in the last layer is taken as the representation vector of this edge, denoted as \kdd{$\vec{h}_{\langle u, v \rangle} \in \mathbb{R}^d$}.

\subsection{Query-Conditioned Graph Modulation} \label{sec.graph-view}
Beyond the edge level, \model\ fuses the edge  representations into a whole-graph representation to facilitate structure matching between query and input graphs.

\stitle{Query graph representation.}
We employ a typical readout function \iclrcite{xu2018powerful,lee2019self,yao2020graph}  on a query graph, by aggregating all edge representations in the query. Given a query graph $\bQ$, its whole-graph representation is computed as
\begin{equation} \label{eq.aggr-query}
    \vec{h}_{\bQ} = \sigma(\vec{Q}\cdot \textsc{Aggr}(\{\vec{h}_{\langle u, v \rangle} | {\langle u, v \rangle} \in E_{\bQ}\})),
\end{equation}
where $\vec{Q}\in \mathbb{R}^{d\times d}$ is a learnable weight matrix shared by all query graphs, and we use sum for the aggregation. Intuitively, the query graph representation simply pools all edge representations together uniformly.

\stitle{Input graph representation.}
To generate a whole-graph representation for the input graph, a straightforward way is to follow Eq.~\iclrref{eq.aggr-query} by regarding all edges uniformly. However, for an input graph, the space of possible query graphs is enormous. In particular, different queries are often characterized by distinct structures, which implies that different parts of the input graph will be triggered to match different queries.
Therefore, aggregating all edges in the input graph uniformly cannot retain sufficiently specific structural properties \wrt~each query. 
In other words, using a fixed whole-graph representation for the input graph cannot tailor to each query well for effective structure matching. Thus, we propose to modulate the input graph conditioned on the query, to adapt the whole-graph representation of the input graph to each query.
To this end, we leverage Feature-wise Linear Modulation (FiLM) \iclrcite{perez2018film,liu2021nodewise,liu2021tail} on the edge representations in the input graph, conditioned on the query, in order to retain query-specific structures. The modulation is essentially a scaling and shifting transformation to adapt the edge representations of the input graph to the query. Given an input graph $\bG$, for each edge $e={\langle u, v \rangle} \in E_\bG$ we modulate its representation $\vec{h}_{\langle u, v \rangle}$ into $\tilde{\vec{h}}_{\langle u, v \rangle}$, as follows.
\begin{equation}
\tilde{\vec{h}}_{\langle u, v \rangle} = (\gamma_{\langle u, v \rangle} + \vec{1}) \odot \vec{h}_{\langle u, v \rangle} + \beta_{\langle u, v \rangle},
\end{equation}
where $\gamma_{\langle u, v \rangle}$ and $\beta_{\langle u, v \rangle}\in\mathbb{R}^{d}$ are FiLM factors for scaling and shifting, respectively, $\odot$ denotes the Hadamard product, and $\vec{1}\in\mathbb{R}^{d}$ is a vector filled with ones to center the scaling factor around one. Note that the FiLM factors $\gamma_{\langle u, v \rangle}$ and $\beta_{\langle u, v \rangle}$ are not directly learnable, but are instead generated by a secondary network \iclrcite{ha2016hypernetworks} conditioned on the original edge representation $\vec{h}_{\langle u, v \rangle}$ and the query representation $\vec{h}_{\bQ}$. More specifically,
\begin{align}
\gamma_{\langle u, v \rangle} = \sigma(\vec{W}_{\gamma}\vec{h}_{\langle u, v \rangle} + \vec{U}_{\gamma}\vec{h}_{\bQ} + \vec{b}_{\gamma}),\\ 
{\beta}_{\langle u, v \rangle} = \sigma(\vec{W}_{\beta}\vec{h}_{\langle u, v \rangle} + \vec{U}_{\beta}\vec{h}_{\bQ} + \vec{b}_{\beta}),
\end{align}
where $\vec{W}_{\gamma}, \vec{U}_{\gamma}, \vec{W}_{\beta}, \vec{U}_{\beta}\in\mathbb{R}^{d\times d}$ are learnable weight matrices, and $\vec{b}_{\gamma}, \vec{b}_{\beta}\in\mathbb{R}^{d}$ are learnable bias vectors. 
The modulated edge representations can be further fused via a readout function, to generate a modulated whole-graph representation for the input graph, which is tailored toward each query to enable more precise matching between the input graph and query. 
Concretely, consider a query graph $\bQ$ and an input graph $\bG$. We formulate the $\bQ$-conditioned representation for $\bG$, denoted  $\vec{h}^{\bQ}_{\bG}\in\mathbb{R}^{d}$, by aggregating the modulated edge representations of $\bG$ in the following.
\begin{equation} \label{eq.aggr-graph}
    \vec{h}^{\bQ}_{\bG} = \sigma(\vec{G}\cdot \textsc{Aggr}(\{\tilde{\vec{h}}_{\langle u, v \rangle}| {\langle u, v \rangle} \in E_{\bG}\})),
\end{equation}
where $\vec{G} \in \mathbb{R}^{d\times d}$ is a learnable weight matrix shared by all input graphs, and we use sum for the aggregation.
\subsection{Counter Module and Overall Objective}
With the whole-graph representations of the query and input graphs, we design a counter module to estimate the count of subgraph isomorphisms, and formulate the overall objective.

\stitle{Counter module.}
We estimate the count of isomorphisms based on the structure matchability between the query and input graph. 
Given the query graph $\bQ$ and input graph $\bG$, we predict the number of subgraphs on $\bG$ which are isomorphic to $\bQ$
by 
\begin{equation} \label{eq.counter}
    \hat{n}(\bQ,\bG) = \textsc{ReLU}(\vec{w}^\top\textsc{Match}(\vec{h}_{\bQ}, \vec{h}^{\bQ}_{\bG}) + b),
\end{equation}
where $\textsc{Match}(\cdot,\cdot)$ outputs a $d_m$-dimensional vector to represent the matchability between its arguments, 
and $\vec{w} \in\mathbb{R}^{d_m}, b\in \mathbb{R}$ are the learnable weight vector and bias, respectively.
Here a ReLU activation is used to ensure that the prediction is non-negative. 
Note that $\textsc{Match}(\cdot,\cdot)$ can be any function---we adopt a fully connected layer (FCL) such that
$\textsc{Match}(\vec{x},\vec{y})=\textsc{FCL}(\vec{x}\parallel\vec{y}\parallel \vec{x}-\vec{y}\parallel \vec{x}\odot\vec{y})$.

\stitle{Overall objective.}
Based on the counter module, we formulate the overall training loss.
Assume a set of training triples $\bT=\{(\bQ_i, \bG_i, n_i)\mid i=1,2,\ldots\}$, where $n_i$ is the ground truth count for query $\bQ_i$ and input graph $\bG_i$. The ground truth can be evaluated by classical exact algorithms \iclrcite{cordella2004sub}. 
Subsequently, we minimize the following loss: 
\begin{equation} \label{eq.obj}
       \frac{1}{|\bT|}\sum_{(\bQ_i, \bG_i, n_i)\in\bT} |\hat{n}(\bQ_i,\bG_i)-n_i| + \lambda\bL_{\text{FiLM}} + \mu \|\Theta\|_2^2,
\end{equation}
where $\bL_{\text{FiLM}}$ is a regularizer on the FiLM factors and $\|\Theta\|_2^2$ is a L2 regularizer on the model parameters, and $\lambda,\mu$ are hyperparameters to control the weight of the regularizers. Specifically, the FiLM regularizer is designed to smooth the modulations to reduce overfitting, by encouraging less scaling and shifting as follows.
\begin{equation}
  \hspace{-1mm}  \bL_{\text{FiLM}}=\sum_{(\bQ_i, \bG_i, n_i)\in\bT}\sum_{{\langle u, v \rangle}\in E_{\bG_i}}\|\gamma_{{\langle u, v \rangle}}\|^2_2 + \|\beta_{{\langle u, v \rangle}}\|^2_2.  \hspace{-1mm}
\end{equation}
We also present the training algorithm and a complexity analysis in Appendix~A.
\subsection{Theoretical Analysis of \model}
The proposed \model\ capitalizes on edge-centric message passing, which is fundamentally more powerful than conventional node-centric counterparts.
This conclusion can be theoretically shown by the below lemma and theorem. 
\begin{lemma}[Generalization] \label{theo.connection}
\model\ can be reduced to a node-centric GNN, \ie, \model\ can be regarded as a generalization of the latter. \qed
\end{lemma}
In short, \model\ can be reduced to a node-centric GNN by removing some input information and merging some edge representations. This demonstrates that \model\ is at least as powerful as the node-centric GNNs. We present the proof of Lemma~\ref{theo.connection} in Appendix~B.
\begin{theorem}[Expressiveness] \label{theo.comparison}
\model\ is more powerful than node-centric GNNs, which means (i) for any two non-isomorphic graphs that can be distinguished by a node-centric GNN, they can also be distinguished by \model; and (ii) there exists two non-isomorphic graphs that can be distinguished by \model\ but not by a node-centric GNN. \qed
\end{theorem}
Intuitively, edge-centric GNNs are capable of capturing fine-grained structural information, as any node can be viewed as a collapse of edges around the node. Therefore, by treating edges as the first-class citizens, \model\ becomes more powerful. 
The proof of Theorem~\ref{theo.comparison} can be found in Appendix~B.

\begin{table}[tbp]
\center
\small
\caption{Summary of datasets.\label{table.datasets}}
\resizebox{0.85\columnwidth}{!}{%
\begin{tabular}{@{}c|rrrr@{}}
\toprule
	& SMALL & LARGE & MUTAG & OGB-PPA \\
\midrule
     \# Queries &75 & 122 & 24 &12 \\
     \# Graphs & 6,790 & 3,240 & 188 & 6,000\\
     \# Triples & 448,140 & 395,280 & 4,512 & 57,940 \\
     Avg($|V_{\bQ}|$) & 5.20 & 8.43 & 3.50 & 4.50 \\
     Avg($|E_{\bQ}|$) & 6.80 & 12.23 & 2.50 & 4.75 \\
     Avg($|V_{\bG}|$) & 32.62 & 239.94 & 17.93 & 152.75 \\
     Avg($|E_{\bG}|$) & 76.34 & 559.68 & 39.58 & 1968.29\\
     Avg(Counts) & 14.83 & 34.42 & 17.76 & 13.83\\
     Max($|L|$) & 16 & 64 & 7 & 8\\
     Max($|L'|$) & 16 & 64 & 4 & 1\\
 \bottomrule
\end{tabular}}
\end{table}

\section{Experiments}
In this section, we empirically evaluate the proposed model \model\
in comparison to the state of the art.

\begin{table*}[tbp] 
    \centering
    \small
     \addtolength{\tabcolsep}{-1mm}
    \caption{Evaluation in the main setting. VF2 generates the exact counts, giving a perfect MAE (0) and Q-error (1). Time refers to the total inference time on all test triples, in seconds. Except VF2, the best method is bolded and the runner-up is underlined. 
    }
    \label{table.main-results}%
    \resizebox{1.0\textwidth}{!}{%
    \begin{tabular}{@{}l|ccc|ccc|ccc|ccc@{}}
    \toprule
   \multirow{2}*{Methods} & \multicolumn{3}{c|}{SMALL} &
   \multicolumn{3}{c|}{LARGE} & \multicolumn{3}{c|}{MUTAG} & \multicolumn{3}{c}{OGB-PPA}  \\ 
      & MAE $\downarrow$ & Q-error $\downarrow$  & Time/s $\downarrow$ & MAE $\downarrow$ & Q-error $\downarrow$  & Time/s $\downarrow$ & MAE $\downarrow$ & Q-error $\downarrow$  & Time/s $\downarrow$  
      & MAE $\downarrow$ & Q-error $\downarrow$  & Time/s $\downarrow$\\\midrule\midrule
    \method{GCN}  & 14.8 $\pm$ 0.5 & 2.1 $\pm$ 0.1 & 7.9 $\pm$ 0.2 & 33.0 $\pm$ 0.4 & 3.5 $\pm$ 1.0 & \underline{29.8} $\pm$ 0.7 & 19.9 $\pm$ 9.7 & 4.2 $\pm$ 1.5 & 0.88 $\pm$ 0.02
    & 36.8 $\pm$ 1.4 & 2.1 $\pm$ 0.4 & 12.5 $\pm$ 0.3\\
    \method{GraphSAGE}  & 14.0 $\pm$ 2.7 & 2.5 $\pm$ 0.8 & \textbf{7.0} $\pm$ 0.1 & 33.8 $\pm$ 1.6 & \underline{3.1} $\pm$ 0.4 & \textbf{27.5} $\pm$ 1.3 & 13.9 $\pm$ 2.8 & 4.7 $\pm$ 0.8 & 0.88 $\pm$ 0.02 
    & 32.5 $\pm$ 4.5 & 2.5 $\pm$ 0.5 & \textbf{11.1} $\pm$ 0.1\\
     \method{GAT}  & 12.2 $\pm$ 0.7 & 2.0 $\pm$ 0.5 & 14.3 $\pm$ 0.3 & 37.3 $\pm$ 5.2 & 6.0 $\pm$ 1.2 & 59.4 $\pm$ 0.7 & 30.8 $\pm$ 6.7 & 6.0 $\pm$ 0.3 & 0.91 $\pm$ 0.01
     & 35.8 $\pm$ 2.4 & 2.2 $\pm$ 0.6 & 30.4 $\pm$ 0.8\\
    \method{DPGCNN}  & 16.8 $\pm$ 0.7 & 2.9 $\pm$ 0.2 & 21.7 $\pm$ 0.4 
    & 39.8 $\pm$ 3.7 & 5.4 $\pm$ 1.6 & 64.8 $\pm$ 0.9 
    & 27.5 $\pm$ 2.5 & 4.9 $\pm$ 0.6 & 1.54 $\pm$ 0.01
    & 38.4 $\pm$ 1.2 & 2.3 $\pm$ 0.3 & 19.4 $\pm$ 0.7\\
    \method{DiffPool}  & 14.8 $\pm$ 2.6 & 2.1 $\pm$ 0.4 & \textbf{7.0} $\pm$ 0.1 & 34.9 $\pm$ 1.4 & 3.8 $\pm$ 0.7 & 32.5 $\pm$ 0.7 & 6.4 $\pm$ 0.3 & 2.5 $\pm$ 0.2 & 0.86 $\pm$ 0.00  
    & 35.9 $\pm$ 4.7 & 2.7 $\pm$ 0.3 & 15.4 $\pm$ 2.2\\
    \method{GIN}  & 12.6 $\pm$ 0.5 & 2.1 $\pm$ 0.1 & \underline{7.1} $\pm$ 0.0 & 35.9 $\pm$ 0.6 & 4.8 $\pm$ 0.2 & 33.5 $\pm$ 0.6 & 21.3 $\pm$ 1.0 & 5.6 $\pm$ 0.7 & 0.41 $\pm$ 0.01
    & 34.6 $\pm$ 1.4 & 2.5 $\pm$ 0.5 & \underline{12.3} $\pm$ 0.4\\\midrule
    \method{RGCN-Sum}  & 24.2 $\pm$ 6.1 & 3.7 $\pm$ 1.2 & 13.2 $\pm$ 0.1 & 80.9 $\pm$ 26.3 & 6.3 $\pm$ 1.3 & 61.8 $\pm$ 0.2 & 8.0 $\pm$ 0.8 & \textbf{1.5} $\pm$ 0.1 & 0.89 $\pm$ 0.01
    & 34.5 $\pm$ 13.6 & 4.7 $\pm$ 0.8 & 33.0 $\pm$ 0.2\\
    \method{RGCN-DN}  & 16.6 $\pm$ 2.3 & 3.2 $\pm$ 1.3 & 48.1 $\pm$ 0.2 & 73.7 $\pm$ 29.2 & 9.1 $\pm$ 4.2 & 105.0 $\pm$ 0.4 & 7.3 $\pm$ 0.8 & 2.6 $\pm$ 0.2 & 1.19 $\pm$ 0.04
    & 57.1 $\pm$ 15.7 & 5.0 $\pm$ 1.3 & 31.2 $\pm$ 0.1\\
    \method{RGIN-Sum} & 10.7 $\pm$ 0.3 & 2.0 $\pm$ 0.2 & 12.2 $\pm$ 0.0 & 33.2 $\pm$ 2.2 & 4.2 $\pm$ 1.3 & 61.4 $\pm$ 1.0 & 10.8 $\pm$ 0.9 & 1.9 $\pm$ 0.1 & 0.45 $\pm$ 0.02
    & 29.1 $\pm$ 1.7 & 1.2 $\pm$ 0.6 & 21.0 $\pm$ 1.2\\
    \method{RGIN-DN} & 11.6 $\pm$ 0.2 & 2.4 $\pm$ 0.0 & 49.7 $\pm$ 1.8 & 32.5 $\pm$ 1.9 & 4.3 $\pm$ 2.0 & 104.0 $\pm$ 1.5 & 8.6 $\pm$ 1.9 & 3.3 $\pm$ 0.8 & 0.73 $\pm$ 0.03
    & 35.8 $\pm$ 6.4 & 4.4 $\pm$ 1.1 & 28.8 $\pm$ 0.3\\
    \method{DMPNN-LRP} & \underline{9.1} $\pm$ 0.2 & \underline{1.5} $\pm$ 0.1 & 32.4 $\pm$ 1.4 & \textbf{28.1} $\pm$ 1.3 & 3.4 $\pm$ 1.5 & 184.2 $\pm$ 1.8 & \underline{5.4} $\pm$ 1.8 & \underline{1.8} $\pm$ 1.0 & \underline{0.13} $\pm$ 0.05
    & \textbf{25.6} $\pm$ 4.9 & \underline{1.1} $\pm$ 1.3 & 63.0 $\pm$ 0.6\\
    \midrule
    \method{\model} & \textbf{8.5} $\pm$ 0.0 & \textbf{1.4} $\pm$ 0.1 & 7.9 $\pm$ 0.3 & \underline{30.9} $\pm$ 4.3 & \textbf{2.5} $\pm$ 0.5 & 59.2 $\pm$ 1.7 & \textbf{4.2} $\pm$ 0.1 &
    \underline{1.8} $\pm$ 0.0 & \textbf{0.02} $\pm$ 0.00 
    & \underline{28.7} $\pm$ 3.9 & \textbf{1.0} $\pm$ 0.2 & 18.1 $\pm$ 0.6\\\midrule
    \method{VF2} & 0 & 1 & 1049.2 $\pm$ 2.7 & 0 & 1 & 9270.5 $\pm$ 5.9 & 0 & 1 & 1.30 $\pm$ 0.04 & 0 & 1 & 5836.3 $\pm$ 4.8 \\
    \method{Peregrine} & - & - & 72.4 $\pm$ 2.0 & - & - & 904.2 $\pm$ 4.5 & - & - & 0.2 $\pm$ 0.03 & - & - & 450.1 $\pm$ 3.9 \\
    \bottomrule
        \end{tabular}}
\end{table*}

\subsection{Experimental Setup}

\stitle{Datasets.}
We conduct the evaluation on four datasets shown in Table~\ref{table.datasets}.
In particular, \emph{SMALL} and \emph{LARGE} are two synthetic datasets, which are generated by the query and graph generators presented by a previous study \cite{liu2020neural}. 
On the other hand, \emph{MUTAG} \cite{zhang2018end}  and \emph{OGB-PPA} \cite{hu2020ogb} are two real-world datasets. In particular, MUTAG consists of 188 nitro compound graphs, and OGB-PPA consists of 6,000 protein association graphs. While graphs in MUTAG and OGB-PPA are taken as our input graphs, we use the query generator \cite{liu2020neural} to generate the query graphs. As each dataset consists of multiple query and input graphs, we couple each query graph $\bQ$ with an input graph $\bG$ to form a training triple $(\bQ, \bG, n)$ with $n$ denoting the ground-truth count given by an exact algorithm \method{VF2} \cite{cordella2004sub}.
More details of the datasets are given in Appendix~C.

\stitle{Baselines.}
We compare \model\ with the state-of-the-art approaches in two main categories. 
We provide further details and settings for the baselines in Appendix~D.

(1) \emph{Conventional GNNs}: \method{GCN} \cite{kipf2016semi}, \method{GAT} \cite{velivckovic2017graph}, \method{GraphSAGE} \cite{hamilton2017inductive},  \method{DPGCNN  \cite{monti2018dual},} \method{GIN} \cite{xu2018powerful},  and \method{DiffPool} \cite{ying2018hierarchical}. They capitalize on node-centric message passing, followed by a readout function to obtain the whole-graph representation. Except DiffPool which utilizes a specialized hierarchical readout, we employ a sum pooling over the node representations for the readout in other GNNs.

(2) \emph{GNN-based isomorphism counting models}, including four variants proposed by \cite{liu2020neural}, namely \method{RGCN-DN}, \method{RGCN-Sum}, \method{RGIN-DN}, \method{RGIN-Sum}, as well as \method{LRP} \cite{zhengdao2020can} and \method{DMPNN-LRP} \cite{liu2022graph}, a better variant of DMPNN. They are purposely designed GNNs for subgraph isomorphism counting, relying on different GNNs such as \method{RGCN} \cite{schlichtkrull2018modeling}, \method{RGIN} \cite{xu2018powerful} and local relational pooling \cite{zhengdao2020can}) for node representation learning, followed by a specialized readout suited for isomorphism matching. In particular, the two variants \method{RGCN-DN} and \method{RGIN-DN} utilize \method{DiamNet} \cite{liu2020neural}, whereas \method{RGCN-Sum} and \method{RGIN-Sum} utilize the simple sum-pooling.

Finally, we also include a classical approach \method{VF2} \cite{cordella2004sub} and a state-of-the-art approach \method{Peregrine} \cite{jamshidi2020peregrine}, both of which evaluate the exact counts.
Note that there are many other exact approaches, but they are not suitable baselines due to their lack of generality. In particular, many approaches have algorithmic limitations that can only process small queries, \eg, up to 4 nodes in PGD \cite{ahmed2015efficient}, 5 nodes in RAGE \cite{marcus2012rage} and ORCA \cite{hovcevar2014combinatorial}, and 6 nodes in acc-Motif \cite{meira2014acc}. More broadly speaking, since the counting task is \#P-complete, any exact algorithm is bound to suffer from prohibitively high running time once the queries become just moderately large. Besides, some approaches cannot handle certain kinds of queries or input graphs. For example, SCMD \cite{wang2012symmetry} and PATCOMP \cite{jain2017impact} are not applicable to directed graphs, and PGD \cite{ahmed2015efficient} and RAGE \cite{marcus2012rage} do not handle graphs with node or edge labels. Although Peregrine does not support directed edges or edge labels either, we run it on our datasets by ignoring edge directions/labels if any, and focus on its time cost only. In addition, there are also a number of statistically approximate methods, but they have similar shortcomings. For example, tMotivo and L8Motif \cite{bressan2021faster} can only handle query graphs with no more than 8 or 16 nodes. 

\stitle{Settings and parameters.}
For SMALL and LARGE datasets, we randomly sample 5000 triples for training, 1000 for validation, and the rest for testing. 
For MUTAG, due to its small size, we randomly sample 1000 triples for training, 100 for validation, and the rest for testing. 
For OGB-PPA, we divide the triples into training, validation and testing sets with a proportion of 4:1:5.
We also evaluate the impact of training size on the model performance in Appendix~F. We report further model and parameter settings of \model\ in Appendix~E.

\stitle{Evaluation.}
We employ mean absolute error (MAE) and Q-error \cite{zhao2021learned} to evaluate the effectiveness of \model. The widely used metric MAE measures the magnitude of error in the prediction.
In addition, Q-error measures a relative error defined by $\operatorname{max}(\frac{n}{\hat{n}},\frac{\hat{n}}{n})$, where $n$ denotes the ground-truth count and $\hat{n}$ denotes the predicted count.%
\footnote{If the ground-truth or predicted count is less than 1, we assume a pseudocount of 1 for the calculation of q-error.}
Both metrics are better when smaller: the best MAE is 0 while the best Q-error is 1.
We further report the inference time for all the approaches to evaluate their efficiency on answering queries, as well as their training time.
We repeat all experiments with five runs, and report their average results and standard deviations.

\subsection{Performance Evaluation}
To comprehensively evaluate the performance, we compare \model\ with the baselines in two settings: 
(1) a main setting with triples generated by all the query graphs and input graphs;
(2) a secondary setting with triples generated by all input graphs associated with only one query graph. 
Note that the main setting represents a more general scenario, in which we compare with all baselines except \method{LRP}. However, due to the particular design of \method{LRP} that requires a number of input graphs coupled with one query graph and the corresponding ground-truth count during training, we use the secondary setting only for this baseline. Our model can flexibly work in both settings.

\stitle{Testing with main setting.}
As discussed, we compare \model\ with all baselines except \method{LRP} in this more general scenario, where the triples are generated by coupling every pair of query and input graphs. 

We report the results in Table~\ref{table.main-results}, and compare the aspects of effectiveness and efficiency.
In terms of \emph{effectiveness} measured by MAE and Q-error, \model\ can generally outperform other GNN-based models. 
In the several cases where \model\ is not the best among the GNN-based models, it still emerges as a competitive runner-up. This demonstrates the two key modules of \model, namely, edge-centric aggregation and query-conditioned graph modulation, 
can improve structure matching between input graphs and structurally diverse queries.
In terms of \emph{efficiency} measured by the query time, we make three observations. First, \model\ achieves 65x$\sim$324x speedups over the classical \method{VF2}. While the other exact method Peregrine is orders of magnitude faster than VF2, Count-GNN can still achieve 8x$\sim$26x speedups over Peregrine. Second, \model\ is also more efficient than other GNN-based isomorphism counting models. On the one hand, \model\ achieves 3.1x$\sim$6.5x speedups over DMPNN-LRP, which is generally the second best GNN-based method after \model\ in terms of effectiveness. On the other hand, \model\ also obtains consistent and notable speedups over the fastest \method{RGCN}/\method{RGIN} variant (\ie, \method{RGIN-Sum}), while reducing the errors by 20\% or more in most cases. Finally, although many conventional GNNs can achieve a comparable or faster query time, their efficiency comes at the expense of much worse errors than \model, by at least 30\% in most cases.

Furthermore, we compare the training time of GNN-based approaches in Table~\ref{table.training-time}. Our proposed \model\ generally requires relatively low training time on SMALL and MUTAG, while having comparable training time to the baselines on LARGE and OGB-PPA. 

\begin{table}[tbp]
\centering
\small
    \caption{Comparison of training time per epoch, in seconds.\label{table.training-time}}
        \resizebox{0.8\linewidth}{!}{%
        \begin{tabular}{@{}c|c|c|c|c@{}}
\toprule
	Methods & SMALL & LARGE & MUTAG & OGB-PPA \\\midrule
 \method{GCN}  & 0.8 & 1.1 & 0.35 & 1.0   \\
\method{GraphSAGE}  & 0.8 & 1.0 & 0.35 & 0.9   \\
\method{GAT}  & 1.0 & 2.4 & 0.39 & 1.5  \\
\method{DiffPool}  & 0.8 & 1.3 & 0.34 & 1.2  \\
\method{GIN}  & 0.5 & 1.0 & 0.19 & 0.7 \\ 
 \method{RGCN-SUM}  & 1.0 & 2.4 & 0.38 & 1.7  \\
 \method{RGCN-DN}  & 1.5 & 3.7 & 0.50 & 2.2  \\
 \method{RGIN-SUM}  & 0.7 & 1.9 & 0.23 & 1.3  \\
\method{RGIN-DN}  & 1.4 & 2.9 & 0.39 & 1.8  \\ 
\method{\model}  & 0.4 & 2.5 & 0.04  & 1.3 \\
 \bottomrule
\end{tabular}} 
\end{table}

\stitle{Testing with secondary setting.}
We also generate another group of triples for comparison with the baseline \method{LRP}, in the so-called secondary setting due to the design requirement of \method{LRP}. In particular, we only evaluate on three datasets SMALL, LARGE and MUTAG, as \method{LRP} is out-of-memory on OGB-PPA with large and dense graphs. For each dataset we select three query graphs of different sizes (see Appendix~C). 
On each dataset, we couple each query graph with all the input graphs, thus forming 6790/3240/188 triples for each query in SMALL/LARGE/MUTAG, respectively. Besides, we split the triples of SMALL and LARGE in the ratio of 1:1:2 for training, validation and testing, while using the ratio of 1:1:1 for MUTAG.

The results are reported in Table~\ref{table.results-lrp}. We observe that \model\ consistently outperforms \method{LRP} in terms of effectiveness, significantly reducing MAE by 43\% and Q-error by 64\% on average. This verifies again the power of the two key modules in \model. For efficiency, neither \model\ nor LRP emerges as the clear winner.

\begin{table}[!t] 
    \centering
    \small
    \addtolength{\tabcolsep}{-1mm}
    \caption{Evaluation in the secondary setting. Time refers to the total inference time on all test triples, in seconds. The better method for each query is bolded. 
    }
    \label{table.results-lrp}%
    \resizebox{1\columnwidth}{!}{%
    \begin{tabular}{@{}cl|ccc|ccc|ccc@{}}
    \toprule
   \multicolumn{2}{l|}{} & \multicolumn{3}{c|}{SMALL} &
   \multicolumn{3}{c|}{LARGE} &   \multicolumn{3}{c}{MUTAG}\\
      \multicolumn{2}{l|}{} & MAE  & Q-err  & Time/s  & MAE  & Q-err  & Time/s  & MAE  & Q-err  & Time/s  \\\midrule
     \multirow{2}{*}{$\bQ_1$} & \method{LRP}  & 11.5 & 3.6 & 0.13
     & 126.1 & 38.3 & \textbf{0.04}
     & 12.3 & 2.1 & 0.01  \\
     & \method{\model}  & \textbf{3.0} & \textbf{1.4} & \textbf{0.04} 
     & \textbf{111.2} & \textbf{2.9} & 0.22 
     & \textbf{2.5} & \textbf{1.2} & \textbf{0.00}  \\\midrule
     \multirow{2}{*}{$\bQ_2$} & \method{LRP}  & 12.6 & 4.6 & 0.12
     & 19.8 & 3.7 & \textbf{0.04}
     & 7.8 & 2.9 & 0.01  \\
     & \method{\model}  & \textbf{4.6} & \textbf{1.1} & \textbf{0.05} 
     & \textbf{4.3} & \textbf{1.1} & 0.07
     & \textbf{5.0} & \textbf{2.1} & \textbf{0.01}  \\\midrule
     \multirow{2}{*}{$\bQ_3$} & \method{LRP}  & 31.5 & 4.1 & 0.05 
     & 87.2 & 7.1 & \textbf{0.04}
     & 8.3 & 2.8 & 0.01  \\
     & \method{\model}  & \textbf{23.2} & \textbf{1.3} & \textbf{0.03} 
     & \textbf{58.0} & \textbf{1.8} & 0.08 
     & \textbf{4.3} & \textbf{1.8} & \textbf{0.01}  \\\midrule
     \multirow{2}{*}{Avg} & \method{LRP}  & 18.5 & 4.1 & 0.10 
     & 77.7 & 16.4  & \textbf{0.04} 
     & 9.5 & 2.6 & 0.01   \\
     & \method{\model}  & \textbf{10.3} & \textbf{1.3} & \textbf{0.04} 
     & \textbf{57.8} & \textbf{1.9} & 0.12 
     & \textbf{3.9} & \textbf{1.7} & 0.01  \\

     \bottomrule
     \end{tabular}}
\end{table}

\stitle{Summary.} Compared to existing GNN-based models, \model\ generally achieves significant error reductions with comparable or faster training and inference, advancing the research of GNN-based isomorphism counting.

\subsection{Model Analysis}
\kdd{We further investigate various aspects of \model\ on the three datasets SMALL, LARGE and MUTAG.
}

\begin{table}[!t] 
    \centering
    \small
    \addtolength{\tabcolsep}{-1mm}
    \caption{Ablation study on Count-GNN. 
    }
    \label{table.ablation}%
    \resizebox{0.85\columnwidth}{!}{%
    \begin{tabular}{@{}l|cc|cc|cc@{}}
    \toprule
   \multicolumn{1}{c|}{\multirow{2}{*}{Methods}} & \multicolumn{2}{c|}{SMALL} &
   \multicolumn{2}{c|}{LARGE} &   \multicolumn{2}{c}{MUTAG}\\
      & MAE  & Q-error & MAE  & Q-error   & MAE  & Q-error   \\\midrule
     \method{Count-GNN\textbackslash{E}}  & 11.3 & 2.07 & 33.58
     & 4.96 & 18.63 & 5.92 \\
      \method{Count-GNN\textbackslash{M}}  & 8.66 & 1.46 & \textbf{29.65}
     & 3.34 & 4.41 & 1.82 \\
     \method{\model}  & \textbf{8.54} & \textbf{1.41} & 30.91 
     & \textbf{2.46} & \textbf{4.22} & \textbf{1.76} \\
     \bottomrule
     \end{tabular}}
\end{table}

\stitle{Ablation Study.}
To evaluate the impact of each module in \model, we conduct an ablation study by comparing \model\ with its two degenerate variants: (1) \model$\backslash$E, which replaces the edge-centric aggregation with the node-centric \method{GIN}; (2) \model$\backslash$M, which replaces the query-conditioned modulation with a simple sum-pooling as the readout for the input graph. 

We report the results in Table~\ref{table.ablation}. Not surprisingly, the full model generally outperforms the two variants,
further demonstrating the benefit of edge-centric aggregation and query-conditioned modulation.
Furthermore, \model$\backslash$M is usually better than \model$\backslash$E, which implies that edge-centric aggregation may contribute more to the performance boost, possibly due to its more central role in capturing fine-grained structure information by treating edges as the first-class citizen.

\stitle{Scalability, impact of parameters and training size.} We present these results in Appendix~F due to space limitation.

\section{Conclusions}

In this paper, we proposed a novel model called \model\ to approximately solve subgraph isomorphic counting on labeled graphs. 
In terms of modelling, we designed two key modules for \model, namely, edge-centric message passing and query-conditioned graph modulation, to improve structure matching between the query and input graphs. In terms of theory, we showed that edge-centric message passing is more expressive than its node-centric counterpart.
In terms of empirical results, 
we conducted extensive experiments on several benchmark datasets to demonstrate the effectiveness and efficiency of \model.

\section*{Acknowledgments}
This work was supported in part by the National Key Research and Development Program of China under Grant 2020YFB2103803. Dr. Yuan Fang acknowledges the Lee Kong Chian Fellowship 
awarded by Singapore Management University for the 
support of this work. The authors wish to thank Dr.~Yuchen Li from Singapore Management University for his valuable comments on this work.


\bibliography{references}

\begin{thebibliography}{54}
\providecommand{\natexlab}[1]{#1}

\bibitem[{Ahmed et~al.(2015)Ahmed, Neville, Rossi, and
  Duffield}]{ahmed2015efficient}
Ahmed, N.~K.; Neville, J.; Rossi, R.~A.; and Duffield, N. 2015.
\newblock Efficient graphlet counting for large networks.
\newblock In \emph{IEEE International Conference on Data Mining}, 1--10.

\bibitem[{Alon, Yuster, and Zwick(1995)}]{alon1995color}
Alon, N.; Yuster, R.; and Zwick, U. 1995.
\newblock Color-coding.
\newblock \emph{Journal of the ACM}, 42(4): 844--856.

\bibitem[{Bressan, Leucci, and Panconesi(2021)}]{bressan2021faster}
Bressan, M.; Leucci, S.; and Panconesi, A. 2021.
\newblock Faster motif counting via succinct color coding and adaptive
  sampling.
\newblock \emph{ACM Transactions on Knowledge Discovery from Data}, 15(6):
  1--27.

\bibitem[{Carletti et~al.(2017)Carletti, Foggia, Saggese, and
  Vento}]{carletti2017challenging}
Carletti, V.; Foggia, P.; Saggese, A.; and Vento, M. 2017.
\newblock Challenging the time complexity of exact subgraph isomorphism for
  huge and dense graphs with VF3.
\newblock \emph{IEEE transactions on pattern analysis and machine
  intelligence}, 40(4): 804--818.

\bibitem[{Chung et~al.(2014)Chung, Gulcehre, Cho, and
  Bengio}]{chung2014empirical}
Chung, J.; Gulcehre, C.; Cho, K.; and Bengio, Y. 2014.
\newblock Empirical evaluation of gated recurrent neural networks on sequence
  modeling.
\newblock In \emph{NeurIPS Workshop on Deep Learning}.

\bibitem[{Cordella et~al.(2004)Cordella, Foggia, Sansone, and
  Vento}]{cordella2004sub}
Cordella, L.~P.; Foggia, P.; Sansone, C.; and Vento, M. 2004.
\newblock A (sub) graph isomorphism algorithm for matching large graphs.
\newblock \emph{IEEE Transactions on Pattern Analysis and Machine
  Intelligence}, 26(10): 1367--1372.

\bibitem[{Fang et~al.(2016)Fang, Lin, Zheng, Wu, Chang, and
  Li}]{fang2016semantic}
Fang, Y.; Lin, W.; Zheng, V.~W.; Wu, M.; Chang, K. C.-C.; and Li, X.-L. 2016.
\newblock Semantic proximity search on graphs with metagraph-based learning.
\newblock In \emph{IEEE International Conference on Data Engineering},
  277--288.

\bibitem[{Gong and Cheng(2019)}]{gong2019exploiting}
Gong, L.; and Cheng, Q. 2019.
\newblock Exploiting edge features for graph neural networks.
\newblock In \emph{IEEE/CVF Conference on Computer Vision and Pattern
  Recognition}, 9211--9219.

\bibitem[{Ha, Dai, and Le(2017)}]{ha2016hypernetworks}
Ha, D.; Dai, A.; and Le, Q.~V. 2017.
\newblock Hypernetworks.
\newblock In \emph{International Conference on Learning Representations}.

\bibitem[{Hamilton, Ying, and Leskovec(2017)}]{hamilton2017inductive}
Hamilton, W.; Ying, Z.; and Leskovec, J. 2017.
\newblock Inductive representation learning on large graphs.
\newblock In \emph{International Conference on Neural Information Processing
  Systems}, 1024--1034.

\bibitem[{Han, Lee, and Lee(2013)}]{han2013turboiso}
Han, W.-S.; Lee, J.; and Lee, J.-H. 2013.
\newblock Turbo$_\text{iso}$: towards ultrafast and robust subgraph isomorphism
  search in large graph databases.
\newblock In \emph{ACM SIGMOD International Conference on Management of Data},
  337--348.

\bibitem[{He and Singh(2008)}]{he2008graphs}
He, H.; and Singh, A.~K. 2008.
\newblock Graphs-at-a-time: query language and access methods for graph
  databases.
\newblock In \emph{ACM SIGMOD International Conference on Management of Data},
  405--418.

\bibitem[{He et~al.(2016)He, Zhang, Ren, and Sun}]{he2016deep}
He, K.; Zhang, X.; Ren, S.; and Sun, J. 2016.
\newblock Deep residual learning for image recognition.
\newblock In \emph{IEEE Conference on Computer Vision and Pattern Recognition},
  770--778.

\bibitem[{Ho{\v{c}}evar and Dem{\v{s}}ar(2014)}]{hovcevar2014combinatorial}
Ho{\v{c}}evar, T.; and Dem{\v{s}}ar, J. 2014.
\newblock A combinatorial approach to graphlet counting.
\newblock \emph{Bioinformatics}, 30(4): 559--565.

\bibitem[{Hu et~al.(2020)Hu, Fey, Zitnik, Dong, Ren, Liu, Catasta, and
  Leskovec}]{hu2020ogb}
Hu, W.; Fey, M.; Zitnik, M.; Dong, Y.; Ren, H.; Liu, B.; Catasta, M.; and
  Leskovec, J. 2020.
\newblock Open Graph Benchmark: Datasets for Machine Learning on Graphs.
\newblock \emph{arXiv preprint arXiv:2005.00687}.

\bibitem[{Isufi, Gama, and Ribeiro(2021)}]{isufi2021edgenets}
Isufi, E.; Gama, F.; and Ribeiro, A. 2021.
\newblock EdgeNets: Edge varying graph neural networks.
\newblock \emph{IEEE Transactions on Pattern Analysis and Machine
  Intelligence}.

\bibitem[{Jain et~al.(2017)}]{jain2017impact}
Jain, S.; et~al. 2017.
\newblock Impact of memory space optimization technique on fast network motif
  search algorithm.
\newblock In \emph{International Conference on Computer and Computational
  Sciences}, 559--567.

\bibitem[{Jamshidi, Mahadasa, and Vora(2020)}]{jamshidi2020peregrine}
Jamshidi, K.; Mahadasa, R.; and Vora, K. 2020.
\newblock Peregrine: a pattern-aware graph mining system.
\newblock In \emph{European Conference on Computer Systems}, 1--16.

\bibitem[{Jiang et~al.(2020)Jiang, Zhu, Li, and Ji}]{jiang2020co}
Jiang, X.; Zhu, R.; Li, S.; and Ji, P. 2020.
\newblock Co-embedding of nodes and edges with graph neural networks.
\newblock \emph{IEEE Transactions on Pattern Analysis and Machine
  Intelligence}.

\bibitem[{Kipf and Welling(2017)}]{kipf2016semi}
Kipf, T.~N.; and Welling, M. 2017.
\newblock Semi-supervised classification with graph convolutional networks.
\newblock \emph{International Conference on Learning Representations}.

\bibitem[{LeCun et~al.(1998)LeCun, Bottou, Bengio, and
  Haffner}]{lecun1998gradient}
LeCun, Y.; Bottou, L.; Bengio, Y.; and Haffner, P. 1998.
\newblock Gradient-based learning applied to document recognition.
\newblock \emph{Proceedings of the IEEE}, 86(11): 2278--2324.

\bibitem[{Lee, Lee, and Kang(2019)}]{lee2019self}
Lee, J.; Lee, I.; and Kang, J. 2019.
\newblock Self-attention graph pooling.
\newblock In \emph{International Conference on Machine Learning}, 3734--3743.

\bibitem[{Liu et~al.(2020)Liu, Pan, He, Song, Jiang, and Shang}]{liu2020neural}
Liu, X.; Pan, H.; He, M.; Song, Y.; Jiang, X.; and Shang, L. 2020.
\newblock Neural subgraph isomorphism counting.
\newblock In \emph{ACM SIGKDD International Conference on Knowledge Discovery
  and Data Mining}, 1959--1969.

\bibitem[{Liu and Song(2022)}]{liu2022graph}
Liu, X.; and Song, Y. 2022.
\newblock Graph convolutional networks with dual message passing for subgraph
  isomorphism counting and matching.
\newblock In \emph{AAAI Conference on Artificial Intelligence}, 7594--7602.

\bibitem[{Liu et~al.(2021)Liu, Fang, Liu, and Hoi}]{liu2021nodewise}
Liu, Z.; Fang, Y.; Liu, C.; and Hoi, S.~C. 2021.
\newblock Node-wise Localization of Graph Neural Networks.
\newblock In \emph{International Joint Conference on Artificial Intelligence},
  1520--1526.

\bibitem[{Liu, Nguyen, and Fang(2021)}]{liu2021tail}
Liu, Z.; Nguyen, T.-K.; and Fang, Y. 2021.
\newblock Tail-GNN: Tail-node graph neural networks.
\newblock In \emph{ACM SIGKDD Conference on Knowledge Discovery \& Data
  Mining}, 1109--1119.

\bibitem[{Liu et~al.(2018)Liu, Zheng, Zhao, Yang, Chang, Wu, and
  Ying}]{liu2018subgraph}
Liu, Z.; Zheng, V.~W.; Zhao, Z.; Yang, H.; Chang, K. C.-C.; Wu, M.; and Ying,
  J. 2018.
\newblock Subgraph-augmented path embedding for semantic user search on
  heterogeneous social network.
\newblock In \emph{International Conference on World Wide Web}, 1613--1622.

\bibitem[{Marcus and Shavitt(2012)}]{marcus2012rage}
Marcus, D.; and Shavitt, Y. 2012.
\newblock {RAGE}--a rapid graphlet enumerator for large networks.
\newblock \emph{Computer Networks}, 56(2): 810--819.

\bibitem[{Meira et~al.(2014)Meira, M{\'a}ximo, Fazenda, and
  Da~Concei{\c{c}}ao}]{meira2014acc}
Meira, L.~A.; M{\'a}ximo, V.~R.; Fazenda, {\'A}.~L.; and Da~Concei{\c{c}}ao,
  A.~F. 2014.
\newblock {acc-Motif}: Accelerated network motif detection.
\newblock \emph{IEEE/ACM Transactions on Computational Biology and
  Bioinformatics}, 11(5): 853--862.

\bibitem[{Milo et~al.(2002)Milo, Shen-Orr, Itzkovitz, Kashtan, Chklovskii, and
  Alon}]{milo2002network}
Milo, R.; Shen-Orr, S.; Itzkovitz, S.; Kashtan, N.; Chklovskii, D.; and Alon,
  U. 2002.
\newblock Network motifs: simple building blocks of complex networks.
\newblock \emph{Science}, 298(5594): 824--827.

\bibitem[{Monti, Otness, and Bronstein(2018)}]{monti2018motifnet}
Monti, F.; Otness, K.; and Bronstein, M.~M. 2018.
\newblock {MotifNet}: a motif-based graph convolutional network for directed
  graphs.
\newblock In \emph{IEEE Data Science Workshop}, 225--228.

\bibitem[{Monti et~al.(2018)Monti, Shchur, Bojchevski, Litany, G{\"u}nnemann,
  and Bronstein}]{monti2018dual}
Monti, F.; Shchur, O.; Bojchevski, A.; Litany, O.; G{\"u}nnemann, S.; and
  Bronstein, M.~M. 2018.
\newblock Dual-primal graph convolutional networks.
\newblock \emph{arXiv preprint arXiv:1806.00770}.

\bibitem[{Perez et~al.(2018)Perez, Strub, De~Vries, Dumoulin, and
  Courville}]{perez2018film}
Perez, E.; Strub, F.; De~Vries, H.; Dumoulin, V.; and Courville, A. 2018.
\newblock {FiLM}: Visual reasoning with a general conditioning layer.
\newblock In \emph{AAAI Conference on Artificial Intelligence}, 3942--3951.

\bibitem[{Pinar, Seshadhri, and Vishal(2017)}]{pinar2017escape}
Pinar, A.; Seshadhri, C.; and Vishal, V. 2017.
\newblock Escape: Efficiently counting all 5-vertex subgraphs.
\newblock In \emph{International Conference on World Wide Web}, 1431--1440.

\bibitem[{Sankar, Zhang, and Chang(2019)}]{sankar2019meta}
Sankar, A.; Zhang, X.; and Chang, K. C.-C. 2019.
\newblock {Meta-GNN}: metagraph neural network for semi-supervised learning in
  attributed heterogeneous information networks.
\newblock In \emph{IEEE/ACM International Conference on Advances in Social
  Networks Analysis and Mining}, 137--144.

\bibitem[{Schlichtkrull et~al.(2018)Schlichtkrull, Kipf, Bloem, Van Den~Berg,
  Titov, and Welling}]{schlichtkrull2018modeling}
Schlichtkrull, M.; Kipf, T.~N.; Bloem, P.; Van Den~Berg, R.; Titov, I.; and
  Welling, M. 2018.
\newblock Modeling relational data with graph convolutional networks.
\newblock In \emph{European Semantic Web Conference}, 593--607.

\bibitem[{Sun et~al.(2011)Sun, Han, Yan, Yu, and Wu}]{sun2011pathsim}
Sun, Y.; Han, J.; Yan, X.; Yu, P.~S.; and Wu, T. 2011.
\newblock {PathSim}: Meta path-based top-k similarity search in heterogeneous
  information networks.
\newblock \emph{Proceedings of the VLDB Endowment}, 4(11): 992--1003.

\bibitem[{Teixeira et~al.(2018)Teixeira, Cotta, Ribeiro, and
  Meira}]{teixeira2018graph}
Teixeira, C.~H.; Cotta, L.; Ribeiro, B.; and Meira, W. 2018.
\newblock Graph pattern mining and learning through user-defined relations.
\newblock In \emph{IEEE International Conference on Data Mining}, 1266--1271.

\bibitem[{Teixeira et~al.(2020)Teixeira, Kakodkar, Dias, Meira~Jr, and
  Ribeiro}]{teixeira2020sequential}
Teixeira, C.~H.; Kakodkar, M.; Dias, V.; Meira~Jr, W.; and Ribeiro, B. 2020.
\newblock Sequential Stratified Regeneration: MCMC for Large State Spaces with
  an Application to Subgraph Count Estimation.
\newblock \emph{arXiv preprint arXiv:2012.03879}.

\bibitem[{Ullmann(1976)}]{ullmann1976algorithm}
Ullmann, J.~R. 1976.
\newblock An algorithm for subgraph isomorphism.
\newblock \emph{Journal of the ACM}, 23(1): 31--42.

\bibitem[{Veli{\v{c}}kovi{\'c} et~al.(2018)Veli{\v{c}}kovi{\'c}, Cucurull,
  Casanova, Romero, Lio, and Bengio}]{velivckovic2017graph}
Veli{\v{c}}kovi{\'c}, P.; Cucurull, G.; Casanova, A.; Romero, A.; Lio, P.; and
  Bengio, Y. 2018.
\newblock Graph attention networks.
\newblock In \emph{International Conference on Learning Representations}.

\bibitem[{Wang et~al.(2012)Wang, Huang, Wu, and Pan}]{wang2012symmetry}
Wang, J.; Huang, Y.; Wu, F.-X.; and Pan, Y. 2012.
\newblock Symmetry compression method for discovering network motifs.
\newblock \emph{IEEE/ACM Transactions on Computational Biology and
  Bioinformatics}, 9(6): 1776--1789.

\bibitem[{Wang et~al.(2014)Wang, Lui, Ribeiro, Towsley, Zhao, and
  Guan}]{wang2014efficiently}
Wang, P.; Lui, J.~C.; Ribeiro, B.; Towsley, D.; Zhao, J.; and Guan, X. 2014.
\newblock Efficiently estimating motif statistics of large networks.
\newblock \emph{ACM Transactions on Knowledge Discovery from Data}, 9(2):
  1--27.

\bibitem[{Wang et~al.(2019)Wang, Ji, Shi, Wang, Ye, Cui, and
  Yu}]{wang2019heterogeneous}
Wang, X.; Ji, H.; Shi, C.; Wang, B.; Ye, Y.; Cui, P.; and Yu, P.~S. 2019.
\newblock Heterogeneous graph attention network.
\newblock In \emph{The ACM Web Conference}, 2022--2032.

\bibitem[{Wu et~al.(2020)Wu, Pan, Chen, Long, Zhang, and
  Philip}]{wu2020comprehensive}
Wu, Z.; Pan, S.; Chen, F.; Long, G.; Zhang, C.; and Philip, S.~Y. 2020.
\newblock A comprehensive survey on graph neural networks.
\newblock \emph{IEEE Transactions on Neural Networks and Learning Systems}.

\bibitem[{Xia, Li, and Li(2022)}]{xia2022substructure}
Xia, W.; Li, Y.; and Li, S. 2022.
\newblock On the substructure countability of graph neural networks.
\newblock \emph{IEEE Transactions on Knowledge and Data Engineering}.

\bibitem[{Xu et~al.(2019)Xu, Hu, Leskovec, and Jegelka}]{xu2018powerful}
Xu, K.; Hu, W.; Leskovec, J.; and Jegelka, S. 2019.
\newblock How powerful are graph neural networks?
\newblock In \emph{International Conference on Learning Representations}.

\bibitem[{Yan, Yu, and Han(2004)}]{yan2004graph}
Yan, X.; Yu, P.~S.; and Han, J. 2004.
\newblock Graph indexing: A frequent structure-based approach.
\newblock In \emph{ACM SIGMOD International Conference on Management of Data},
  335--346.

\bibitem[{Yang and Li(2020)}]{yang2020nenn}
Yang, Y.; and Li, D. 2020.
\newblock Nenn: Incorporate node and edge features in graph neural networks.
\newblock In \emph{Asian Conference on Machine Learning}, 593--608.

\bibitem[{Yao et~al.(2020)Yao, Zhang, Wei, Jiang, Wang, Huang, Chawla, and
  Li}]{yao2020graph}
Yao, H.; Zhang, C.; Wei, Y.; Jiang, M.; Wang, S.; Huang, J.; Chawla, N.; and
  Li, Z. 2020.
\newblock Graph few-shot learning via knowledge transfer.
\newblock In \emph{AAAI Conference on Artificial Intelligence}, 6656--6663.

\bibitem[{Ying et~al.(2018)Ying, You, Morris, Ren, Hamilton, and
  Leskovec}]{ying2018hierarchical}
Ying, Z.; You, J.; Morris, C.; Ren, X.; Hamilton, W.; and Leskovec, J. 2018.
\newblock Hierarchical Graph Representation Learning with Differentiable
  Pooling.
\newblock In \emph{International Conference on Neural Information Processing
  Systems}, 4800--4810.

\bibitem[{Zhang et~al.(2018)Zhang, Cui, Neumann, and Chen}]{zhang2018end}
Zhang, M.; Cui, Z.; Neumann, M.; and Chen, Y. 2018.
\newblock An end-to-end deep learning architecture for graph classification.
\newblock In \emph{AAAI Conference on Artificial Intelligence}, 4438--4445.

\bibitem[{Zhao et~al.(2021)Zhao, Yu, Zhang, Li, and Rong}]{zhao2021learned}
Zhao, K.; Yu, J.~X.; Zhang, H.; Li, Q.; and Rong, Y. 2021.
\newblock A Learned Sketch for Subgraph Counting.
\newblock In \emph{ACM International Conference on Management of Data},
  2142--2155.

\bibitem[{Zhengdao et~al.(2020)Zhengdao, Lei, Soledad, and
  Bruna}]{zhengdao2020can}
Zhengdao, C.; Lei, C.; Soledad, V.; and Bruna, J. 2020.
\newblock Can Graph Neural Networks Count Substructures?
\newblock In \emph{International Conference on Neural Information Processing
  Systems}, 10383--10395.

\end{thebibliography}
\end{document}